\documentclass[runningheads]{llncs}
\usepackage{graphicx}
\usepackage{cleveref}

\begin{document}

\title{AI Hazard Management: A framework for the 
systematic management of root causes for AI risks}

\titlerunning{AI Hazard Management}

\author{Ronald Schnitzer\inst{1,2} \and
Andreas Hapfelmeier\inst{1} \and
Sven Gaube\inst{1} \and
Sonja Zillner\inst{1,2}}

\authorrunning{R. Schnitzer et al.}
%
\institute{Siemens AG, Munich/Nuremberg, Germany 
\
\email{\{firstname.lastname\}@siemens.com} \and
Technical University of Munich, Munich, Germany 
\email{\{firstname.lastname\}@tum.de}
}

\maketitle              
\begin{abstract}
Recent advancements in the field of Artificial Intelligence (AI) establish the basis to address challenging tasks.
However, with the integration of AI, new risks arise. Therefore, to benefit from its advantages, it is essential to adequately handle the risks associated with AI.
Existing risk management processes in related fields, such as software systems, need to sufficiently consider the specifics of AI. A key challenge is to systematically and transparently identify and address AI risks' root causes - also called AI hazards.
This paper introduces the AI Hazard Management (AIHM) framework, which provides a structured process to systematically identify, assess, and treat AI hazards. The proposed process is conducted in parallel with the development to ensure that any AI hazard is captured at the earliest possible stage of the AI system's life cycle.
In addition, to ensure the AI system's auditability, the proposed framework systematically documents evidence that the potential impact of identified AI hazards could be reduced to a tolerable level. The framework builds upon an AI hazard list from a comprehensive state-of-the-art analysis. Also, we provide a taxonomy that supports the optimal treatment of the identified AI hazards.
Additionally, we illustrate how the AIHM framework can increase the overall quality of a power grid AI use case by systematically reducing the impact of identified hazards to an acceptable level.

\keywords{Trustworthy AI  \and AI risk management \and AI development}
\end{abstract}

\section{Introduction}\label{sec: Intro}

Recent advancements in the field of Artificial Intelligence (AI) establish the basis for solving challenging tasks.
However, despite many opportunities, using AI-based solutions introduces new risks.
For instance, in the field of computer vision, it has been shown that minimal changes in the pixels of an image, so small that a human eye cannot recognize it, can yield an AI classification system to deliver inconsistent results \cite{goodfellow_explaining_2015}.
In medical diagnosis support, this characteristic of AI can have massively negative consequences \cite{apostolidis_survey_2021}.

To fully benefit from the advantages of AI, it is important to handle the risks associated with AI systems adequately. 
To address the problem of how to manage AI risks, the particularities of AI systems need to be reflected.
Well-known AI-related issues include opaqueness, unpredictability, high dependence on data, or vulnerability against security threats of AI systems.
Such pitfalls can negatively affect trustworthiness requirements, including the safety, privacy, and fundamental rights of affected persons, and are, therefore, vital factors on AI risks. Addressing these issues is required to ensure AI systems are developed and operated in compliance with future regulations, such as the Artificial Intelligence Act (AIA) - published by the European Commission \cite{european_commission_proposal_2021}. 
The proposed regulation requires a comprehensive risk management process for all critical AI systems, including safety-critical applications.

To manage AI-related risks, it is required to identify, assess, and treat their corresponding root causes - further also called AI hazards - in a systematic way.
This is a challenging task because AI hazards can appear at different stages of the AI system's lifecycle, manifest at different components of the AI system, and might require various actions.
To illustrate this complexity, consider the example of using deep neural networks (DNNs) in autonomous driving to identify pedestrians.
A specific AI risk in this domain is that the AI system may not be able to detect all types of pedestrians with equal reliability. The authors in \cite{wilson_predictive_2019} present evidence that standard object detection models perform significantly better in identifying light-colored pedestrians than dark-colored ones.
The root causes for this AI risk can stem from different components and stages of the AI system's life cycle.
A possible AI hazard is biased training data, which itself can be introduced at the data collection and the data preparation stage. 
Another possible AI hazard may lie within the model itself, possibly introduced by wrong design decisions, including the so-called loss function or performance evaluation metrics.
In this example, providing non-discriminative outputs is an essential objective of the AI system. Hence this should be reflected in the mentioned design choices. Examples of how to encode fairness in relevant design choices are given in  \cite{caton_fairness_2020}.

Furthermore, risk management in general needs to ensure completeness in terms of addressed risk sources. Several approaches, such as failure mode and effect analysis (FMEA) and preliminary hazard analysis (PHA), introduce valuable mechanisms to analyze hazards in a systematic manner. 
While the FMEA investigates potential failure modes of a system, the PHA relies on a so-called preliminary hazard list, including potential, significant hazards associated with a system's design \cite{popov_risk_2021}. As an example of such a hazard list covering a whole sector, the machinery directive provides a list of machinery-related hazards in Annex I \cite{european_parliament_directive_2006}.

In this work, we aim to transfer this idea to the context of AI applications by providing a wide-ranging list of potential AI hazards that may occur in any scenario.
This list serves as a basis for the identification of relevant AI hazards in a specific AI application.

The contribution of this paper is the AI Hazard Management framework that builds upon this AI hazard list.
The AIHM establishes means to systematically and identify, assess, and treat AI hazards. 
To ensure that any AI hazard is handled at the earliest possible stage in the life cycle of an AI system, the AIHM Framework provides a process executed in parallel to the development of AI systems.
In addition, to fulfill the required level of auditability, the process systematically documents evidence that the impact of identified AI hazards has been reduced to a tolerable level if possible. Otherwise, the framework alerts the developer about unavoidable shortcomings of the AI system.  
In addition, we show in this paper along a use case from the field of power grids how the AIHM framework can increase the overall quality of an AI system by systematically reducing the impact of all identified hazards to a tolerable level. Note, that due to the dominance of data-driven AI systems utilizing Machine Learning (ML) techniques, this is also the focus of the AIHM framework.

The paper is structured as follows. In \Cref{sec: RW}, we present an overview of similar approaches toward the management of AI risks and their root causes, followed by \Cref{sec: Definitions} providing detailed definitions and clarification of used terms and concepts. In \Cref{sec: AI Hazard List}, the state-of-the-art regarding AI hazards is comprehensively analyzed, yielding a preliminary AI hazard list together with a taxonomy of AI hazards. Building upon this, we present the AIHM and its corresponding process in \Cref{sec: Met}. In \Cref{sec: Case Study}, we illustrate the framework's effectiveness along an industrial use case from the field of power grids. We conclude the paper's results and show possible future research directions in \Cref{sec: Conclusion}.

\section{Related Work}\label{sec: RW}
In this section, we provide an overview of the relevant literature concerning the management of AI risks. However, the related work concerning particular AI hazards in detail is described in \Cref{sec: AI Hazard List}.

International standards, such as the ISO IEC Guide 51 \cite{iso_51} or ISO 31000 \cite{iso_31000}, provide general guidelines on managing risks independent of the application domain or used technology.
In addition, initial efforts exist toward standardizing the management of AI risks. For instance, there is the AI Risk Management Framework published by NIST \cite{tabassi_ai_2023}. Additionally, the ISO/IEC JTC SC 42 is currently working on several standards in the field of AI. This includes documents providing general guidelines on the management of risks associated with AI \cite{iso_23894_riskmanagement,iso_42001}. However, these guidelines stay on a relatively abstract level, which makes them challenging to operationalize.

Also, controlling risks is of essential importance in safety-critical applications, particularly in the context of AI. To address safety-related AI risks, safety assurance cases provide a structured argumentation for sufficiently mitigating these risks.
A specific framework for an AI safety assurance case has been proposed in \cite{hawkins_guidance_2021}. Several works also describe the general concepts required for an AI safety assurance case, see e.g. \cite{fingscheidt_safety_2022,mcdermid_towards_nodate,schwalbe_survey_nodate}.
However, while safety assurance cases provide a comprehensive approach to addressing safety-related risks, they do not fully cover other types of risk, including legal, economic, or reputational risks.

The topic of AI risks is also covered by the literature on trustworthy AI.
The concepts of trustworthy AI are built upon the works published by the European High-Level Expert Group (HLEG), see  \cite{european_commission_assessment_2020,european_commission_ethics_2019}.
Although not explicitly addressing AI risks, there is a direct connection between the principles of trustworthy AI and the adequate handling of AI risks.
This is also reflected in the proposed European AI Act. 
Building upon the principles of trustworthy AI, the proposed regulation requires an AI risk management process for every AI system to be classified as "high-risk" by the regulation.
Therefore, approaches assessing the trustworthiness of AI systems provide implicitly valuable insight on how to manage AI risks, see e.g. \cite{floridi_capai_2022,poretschkin_leitfaden_nodate}.
With this work, we aim to build upon all mentioned approaches and propose an operationalizable framework to manage AI risks.

\section{Definitions and background information}
\label{sec: Definitions}
To set the stage for the forthcoming discussion, it is essential to clarify the concept of an AI hazard and offer the necessary context for related terms and definitions. Also, these terms vary in their interpretation, depending on the context and domain of use. However, due to a lack of published standards in the field of AI, a common interpretation of these terms is still to be established. Therefore, it is important to make clear how we interpret them in the scope of this work. In this section, we first provide all normative references and consequently declare our interpretation in the context of AI. 

The first important term is \textit{AI system}. For a definition, we refer to the proposed AI Act, which defines an AI system as: "software that is developed with one or more of the techniques and approaches listed in Annex I and can, for a given set of human-defined objectives, generate outputs such as content, predictions, recommendations, or decisions influencing the environments they interact with" \cite{european_commission_proposal_2021}. Annex I of the AIA provides a list of technologies to be considered AI, which the European Council proposed to narrow down to machine learning and/or logic and knowledge-based approaches.
Note that both, the regulation itself and also the definition of AI system in the AIA, are still under discussion at the time of writing.

While the AIA definition of an AI system restricts its focus to the software aspects, AI systems might be integrated into a larger context, including the system hardware and even the operational environment. We call this larger context \textit{AI application} in the following.
Considering, for instance, an autonomous driving car, the physical vehicle and the operating environment determine the AI application's scope.
The AI system refers to the software that is responsible for fulfilling the human-defined objectives, which in this case can be aggregated to "operate safely in the operational domain".
This might include different \textit{AI models}, such as lane segmentation, and pedestrian detection.

For the scope of this work, it is also important to clarify the terms \textit{risk} and \textit{hazard}, as well as their relationship and connection to AI systems. According to the ISO IEC Guide 51, risk can be understood as a function of the probability and severity associated with a specific hazard \cite{iso_51}. The term \textit{hazard}, as defined by the standard, refers to a potential source of harm \cite{iso_51}.
While this definition is rather abstract, sector-specific standards adapt the definition to their specific requirements. 
For instance, in the road vehicle domain ISO 26262-1, provides a more tangible version, describing a \textit{hazard} as a potential source of harm caused by the behavioral malfunction of an item regarding its intended design \cite{iso_26262}.
For the road vehicle domain, this definition of hazard is appropriate because usually risk comes from non-properly functioning subsystems of the vehicle.

Nonetheless, as previously noted, these definitions are not immediately suitable to apply to AI systems. With the growing adoption of AI-based solutions in the automotive industry, the SOTIF standard (ISO PAS 21448 \cite{iso_sotif}) was introduced as an extension to ISO 26262. The SOTIF methodology defines the desired behavior of the AI system as \textit{intended functionality} and aims to minimize the operational scenarios in which this functionality cannot be assured \cite{iso_sotif}. The specific limitations of the ML algorithms, including their capability “to handle possible scenarios or non-deterministic behavior”, typically fits in the scope of the SOTIF \cite{jenn_identifying_nodate}.



Inspired by these ideas, we aim to extend the concept to AI systems in general. For the road vehicle domain, the intended functionality of an autonomous vehicle is clearly to operate safely in any operational scenario.
To transfer this concept to a more general setting it is important to note that the intended functionality is application dependent.
For example, if an AI system's purpose is to address a classification problem, a violation of the intended functionality at the instance level would be a misclassification. However, the intended functionality might be more complex.
For instance, a classification system may additionally need to maintain similar accuracy for different input groups. An example would be a CV screening tool that should not discriminate against applications based on the gender of the applicant, thus ensuring fairness across all submissions.

Due to the complexity and the fact that AI systems implicitly derive their behavior from large amounts of data, a direct evaluation of the system in its operational state is hindered, presenting one of the main challenges in assessing the risks of AI systems.

Nonetheless, systematic problems persist across the entire lifecycle of AI systems, which can adversely affect an AI system's capacity to maintain the intended functionality.
We designate these issues as \textit{AI hazards} and define them as the underlying causes for the inability to uphold the intended functionality, thereby contributing to AI risks. In essence, AI hazards represent potential sources of harm instigated by AI systems. 


\section{AI Hazard Taxonomy and List}
\label{sec: AI Hazard List}
To efficiently identify, assess and treat AI hazards a systematic approach is required. As a foundation for this, in this section, we introduce a taxonomy for AI hazards. This taxonomy can then be used to understand when, how, and by whom to treat an AI hazard during the development and operation of an AI system. Also, we provide a comprehensive list of AI hazards identified in the literature as well as a corresponding classification according to the proposed taxonomy.
\subsection{AI hazard taxonomy}
\label{subsec: Taxonomy}

AI hazards vary significantly across manifold dimensions, such as the life cycle phase in which they emerge, the system components they impact, and the stakeholders they involve. To ensure systematic management of these AI hazards, we propose a taxonomy that characterizes them based on diverse attributes. This taxonomy subsequently provides an indication of the optimal methods for the detection, evaluation, and mitigation of a specific AI hazard, as well as the responsible parties for executing these actions.

\begin{figure}[h]
    \centering
    \includegraphics[scale = 0.65]{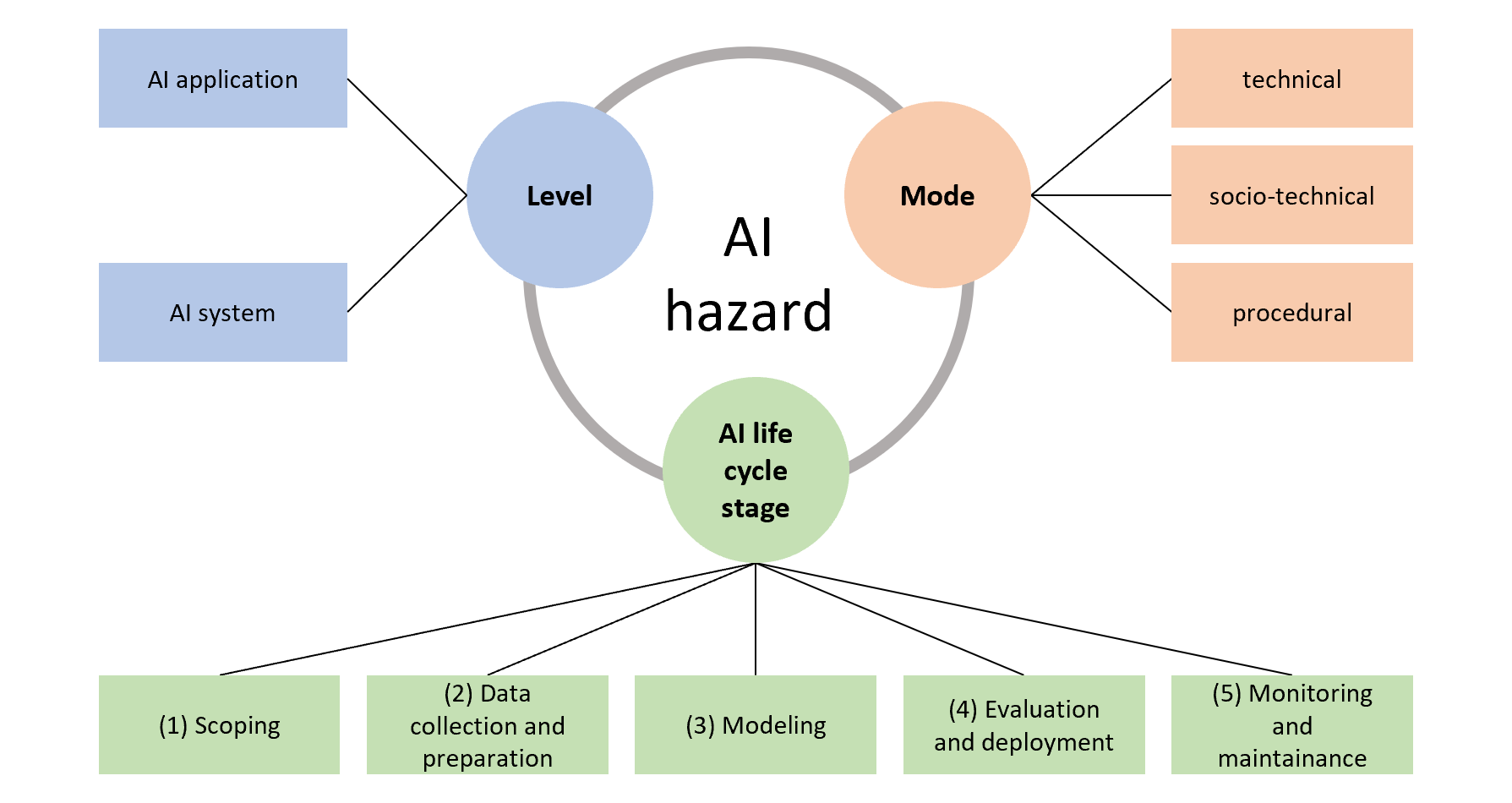}
    \caption{A taxonomy to classify AI hazards along the axes AI hazard level, AI hazard mode, and AI life cycle stage.}
    \label{fig: Taxonomy}
\end{figure}

The taxonomy comprises three distinct axes: AI life cycle stage, AI hazard mode, and AI hazard level. To clarify these axes, we provide a detailed explanation and illustrative examples for each constituent in the subsequent discussion. An overview of the taxonomy is given in \Cref{fig: Taxonomy}.

\textbf{1) AI life cycle - When to address AI hazards} The first axis pertains to the life cycle of the AI system, as AI hazards may materialize during various phases of an AI system's life cycle.
For instance, issues triggered by bias in training data emerge during the data collection and preparation stages. On the other hand, data drift serves as an example of an AI hazard that arises during the AI system's operation. Additionally, certain AI hazards may span multiple phases of the AI system, such as "lack of data understanding". This is because a proper understanding of the data by the AI developer is required in the data collection and preparation stage but also in the modeling stage.

To efficiently mitigate the overall risk posed by an AI hazard to the AI system, it is crucial to identify and address these hazards as early as possible during the AI life cycle. 

However, to construct this axis, a life cycle model is required that meaningfully subdivides the system's life into distinct sub-phases. In the literature, several different models exist describing the life cycle of AI systems, see for instance \cite{floridi_capai_2022,iso_5338_lifecycle,studer_towards_2021}.
While representing the same process continuum, these models primarily differ in the granularity of their stages. 
A majority of them possess an initial stage devoted to the planning and scoping of the AI system. Subsequently, the development stage follows, which is further subdivided. Considering the significance of data and the multitude of AI hazards that can manifest in the corresponding stage, we decided to split the development phase into the data preparation and modeling stages. 
Although some of the cited models differentiate between the evaluation and deployment stages, we opted to merge them into a singular stage, as the development of AI systems typically entails continuous evaluation, and emergent AI hazards transpire during the ultimate evaluation preceding deployment. 
Conclusively, the AI life cycle model terminates with the maintenance and monitoring stage, which aligns with the referenced models.
Therefore, the used AI life cycle model used in this paper is given: 
(1) Scoping, (2) data collection and preparation, (3) modeling, (4) evaluation and deployment, (5) monitoring and maintenance.

\textbf{2) AI Hazard mode -  How to address AI hazards} The second axis of the taxonomy pertains to the mode of an AI hazard, which determines with what methods to assess and treat AI hazards. We distinguish among three distinct classes: technological, socio-technological, and procedural.

Technical AI hazards are the root causes of technical deficiencies in the AI system.
An example of such an AI hazard is overfitting, which describes a model's excessive adaptation to the training dataset.
Quantitative methods to assess (metrics) and treat (mitigation means) exist for technical AI hazards, which might be performed automatically.

In case of overfitting, metrics are based on the comparison of performance between the training and validation datasets, and mitigation means may include regularization techniques, among others.

In contrast to technical AI hazards, socio-technical hazards also require human input related to social and cultural aspects \cite{tabassi_ai_2023}. Human judgment must be employed when deciding on quantification and treatment methods. 
 For instance, AI hazards concerning discrimination and privacy, which are abstract concepts lacking a uniform technical definition, further complicate a clear quantification of the associated risks. 
 Although quantitative methods exist to assess and treat these AI hazards, they require coordination with social and cultural values \cite{iso_24368_ethicalconcerns}.

The third class encompasses procedural AI hazards.
These pertain to issues arising from processes and actions made by individuals involved in the development process.  
Such hazards are not readily quantifiable and necessitate alternative mitigation strategies.
An example of such an AI hazard would be "poor model design choices," which could be expressed, for instance, through a developer's decision to select an unsuitable AI model for a given problem. 
Due to the challenges in quantifying and mitigating these issues, qualitative approaches must be employed. 
In the case of the aforementioned example, a potential strategy might involve requiring the AI developer to provide a documented rationale for their choice. 

Classifying an AI hazard along this axis yields valuable information concerning the proper evaluation and treatment of the hazard in question.

\textbf{3) AI hazard level - Where to address AI hazards} The third axis of the taxonomy pertains to the level, which differentiates between the AI application and system levels, as they are defined in \Cref{sec: Definitions}.

Allocating an AI hazard to its level helps to determine the level at which an action is required. This consequently sets the basis for who is supposed to act.
For instance, the main person responsible for an AI hazard manifesting on the AI system level would be the AI developer, whereas an AI hazard affecting the whole AI application requires a more diverse group, including domain experts.

\subsection{AI hazard list}
In the following, we provide a list of AI hazards identified in the literature and a classification according to the taxonomy introduced in \Cref{subsec: Taxonomy}. The corresponding literature references for each AI hazard are displayed in \Cref{tab: AI hazard overview}.


To generate a comprehensive list of AI hazards for all types of AI systems, we consider a wide range of stakeholders, such as AI developers and users, industrial associations, standardization bodies, and civil society.  
To reflect that AI systems are used in various fields and accomplish different tasks, the consolidation of a holistic list of AI hazards needs to consider state-of-the-art in different fields and disciplines.
This includes various sectors (e.g. autonomous (driving) systems, healthcare), different tasks (e.g. classification, segmentation, clustering), backgrounds (e.g. safety, ethics, security), and components (e.g. data, algorithms, development process).
In the following, we present an initial list of AI hazards that was consolidated along the risk-related requirements of a concrete industrial AI use case (\Cref{sec: Case Study}) for the purpose of demonstrating the value of the overall AI hazard management approach (\Cref{sec: Met}). 
Due to the focus on the industrial use case, this initial AI hazard list mainly covers DNN-related AI hazards and will be extended in our future work. Also, in future work, we aim to enrich the list by providing information regarding concrete methods on how to manage particular AI hazards.

In the following, we provide the AI hazard list, together with a short description of each AI hazard and the corresponding allocation according to the taxonomy. This is indicated in the format [(AI life cycle steps), Mode, Level]. Hereby the first entry is a set of numbers representing the corresponding AI life cycle steps according to the AI life cycle model described in  \Cref{subsec: Taxonomy}. The second entry represents the AI hazard mode, with abbreviations p = procedural, t = technical, s-t = socio-technical. Finally, the third entry represents the AI hazard level with abbreviations sys = AI system level, app = AI application level.

\noindent
\textbf{AIH 1: Inadequate specification of ODD} [\textit{(1, 2, 4, 5), p, app}] The operational design domain (ODD) is a technical description of the application's operational environment, initially conceptualized for autonomous driving systems. An inadequate specification of the ODD limits essential functions such as testing the learned functionality and out-of-distribution detection.

\noindent
\textbf{AIH 2: Inappropriate degree of automation} \textit{[(1, 4, 5), p, app]} The AI application's degree of automation ranges from no automation to fully autonomous. AI applications with a high degree of automation may exhibit unexpected behaviour and pose risks in terms of their reliability and safety.

\noindent
\textbf{AIH 3: Inadequate planning of performance requirements} \textit{[(1, 3, 4, 5), p, sys]} The expected performance of the AI system should be planned adequately. Hereby, an important aspect is that chosen performance metrics are meaningful for presenting the intended functionality. Otherwise, expectations and safety requirements can be unfulfillable at later life cycle stages.

\noindent
\textbf{AIH 4: Insufficient AI development documentation} [\textit{(1, 2, 3), p, sys}] Throughout the development of an AI system, it is vital to document every decision and action taken. This is not only essential to optimize the development process itself but also required for the auditability of the AI system.

\noindent
\textbf{AIH 5: Inappropriate degree of transparency to end users} \textit{[(1, 4, 5), p, app]} The transparency to end users of the AI system increases the user's trust in the AI application. If not adequately integrated into the design, this might prevent the proper operation and cause potential misuse of the AI application. 

\noindent
\textbf{AIH 6: Missing requirements for the implemented hardware} \textit{[(1, 4, 5), p, app]} The development and operation of an AI system can require significant amounts of (computational) power. If not considered in the hardware selection, this can become an issue in development and operation.

\noindent
\textbf{AIH 7: Choice of untrustworthy data source} [\textit{(2), p, sys}] The choice of a trustworthy data source is a first prerequisite in order to fulfill data quality requirements. This is especially the case if third-party data sources are used to develop the AI system.

\noindent
 \textbf{AIH 8: Lack of data understanding} \textit{[(2,3), p, sys]} The correct understanding of the used data for developing an AI system is a prerequisite to avoid data shortcomings and hinders the development of an AI system which is best suiting for the intended functionality.

\noindent
\textbf{AIH 9: Discriminative data bias} \textit{[(2), s-t, sys]} Discriminative data bias describes the systematic discrimination of groups of persons in the form of data shortcomings, such as distributional representation or incorrectness. Data bias can manifest in the model and lead to unfair decisions if not appropriately treated. Note, that the term bias is often used in other contexts, such as data representation. However, these issues are treated by other AI hazards in this list.

\noindent
 \textbf{AIH 10: Harming users' data privacy} \textit{[(2), s-t, sys]} Modern AI systems rely on large amounts of data. If this includes personal data about individuals, the risk of harming the privacy of persons arises. 

\noindent
\textbf{AIH 11: Incorrect data labels} \textit{[(2), t, sys]} Data labels are essential for any supervised learning algorithm since they preset the result of the learning process. If the correctness of the data labels is not given, the AI system is prevented from learning the ground truth and therefore the intended functionality. 

\noindent
\textbf{AIH 12: Data poisoning} \textit{[(2), t, sys]}  Data poisoning describes an attack in the form of an injection of malicious data into the training set. If not prevented, this attack leads the AI system to learn unintended behavior.

\noindent
\textbf{AIH 13: Insufficient data representation} [(2), t, sys]
The distribution of the data used for training a model should match the operational data´s distribution while consisting of sufficiently many samples. 
An important aspect of matching distributions between training and operational data is that also data which is rarely confronting the AI system in operation is represented in the training data.

\noindent
\textbf{AIH 14: Problems of synthetic data} [(2), t, sys]  In the case of sparse data quantity, the simulation or generation of data is a valid alternative. 
However, it is essential to make sure that the simulated data is sufficiently similar to real data, especially in the way the AI system perceives them. Otherwise, generalization to operational data and reliable operational behavior can not be guaranteed. 

\noindent
\textbf{AIH 15: Inappropriate data splitting} \textit{[(2), p, sys]}   In data-driven AI development, the annotated data set is commonly split into training, validation, and test sets, whereby it is essential that the latter is not used for development but only for evaluation. Using the test set for training manipulates the testing strategy, which is the basis of the system's quality assurance. 

\noindent
 \textbf{AIH 16: Poor model design choices} \textit{[(1,3), p, sys]} The model specifications have significant impact on the functionality of an AI system. The developer making wrong decisions might cause the AI system to behave biased and unreliable.

\noindent
\textbf{AIH 17: Over- and underfitting} \textit{[(3), t, sys]}   Over- and underfitting describe the over or insufficient adaption of a model to training data. Both phenomena can cause an AI system to behave unreliably if confronted with operational data. 

\noindent
\textbf{AIH 18: Lack of explainability} \textit{[(3), s-t, sys]}  The explainability of AI systems based on so-called black-box models is often limited. This opaqueness of AI systems can prevent developers from detecting shortcomings in the data or the model itself and decrease the performance and safety levels of the AI system. 

\noindent
 \textbf{AIH 19: Unreliability in corner cases} [\textit{(3), t, sys}] AI systems tend to show unreliable behavior when confronted with rare or ambiguous input data, also called corner cases. Therefore, the controlled behavior is required whenever the AI system is faces a corner case. 

\noindent
\textbf{AIH 20: Lack of robustness} [\textit{(3), t, sys}]  Robustness characterizes the resilience of an AI system's output against minor changes in the input domain. 
A great variation in an AI system's response to small input changes indicates unreliable outputs. 

\noindent
 \textbf{AIH 21: Uncertainty concerns} [\textit{(3), t, sys}]  AI systems should be able not only to return output for a given instance but also to provide a corresponding level of confidence. If such a method is not implemented or not working correctly, this can have a negative impact on performance and safety.   



\noindent
\textbf{AIH 22: Operational data issues} [\textit{(4), t, app}]  Until the deployment of the AI application into its operational environment, the AI system has been tested with a test set that aims to approximate the distribution of operational data. However, an unexpected deviation in this approximation can cause an AI application to behave unreliably. Therefore, its behavior under confrontation with operational data needs to be evaluated.

\noindent
\textbf{AIH 23: Data drift} [\textit{(5), t, sys}]  Data drift is a phenomenon in that distribution of operational input data departs from those used during training. This can cause a degradation in performance.

\noindent
 \textbf{AIH 24: Concept drift} [\textit{(5), t, sys}]  Concept drift refers to a change in the relationship between input variables and model output. If not treated appropriately, concept drift can reduce the reliability of AI systems.

 \begin{table}
\centering
 \caption{An overview of the AI hazards (AIH) together with their related literature.}\label{tab: AI hazard overview}
 \hrule
 \vspace{0.4 cm}
\begin{tabular}{ |p{1cm}|p{1.5cm}||p{1.2cm}|p{1.2cm}||p{1.2cm}|p{1.8cm}||p{1.2cm}|p{2cm}|}

 \hline
 AIH  & sources & AIH & sources & AIH & sources & AIH & sources  \\
 \hline
\hline
 AIH 1 &  \cite{iso_5469_aisafety,willers_safety_2020} &  
 AIH 7 & \cite{iso_29119_softwaretesting} &  
 AIH 13 & \cite{goos_establishing_2003,willers_safety_2020} &
 AIH 19 & \cite{houben_inspect_2022,ouyang_corner_2021,willers_safety_2020}\\

 \hline
 AIH 2 & \cite{iso_5469_aisafety,steimers_sources_2022}  & 
 AIH 8 & \cite{iso_24027_bias,polyzotis_data_2018} &
 AIH 14 & \cite{assefa_generating_2020,sikos_seven_2020} &
  AIH 20 & \cite{houben_inspect_2022,iso_5469_aisafety,steimers_sources_2022,willers_safety_2020} \\

 \hline
 AIH 3 & \cite{steimers_sources_2022,winter_trusted_2021} &  
 AIH 9 & \cite{mehrabi_survey_2021,steimers_sources_2022}   & 
 AIH 15 & \cite{willers_safety_2020,winter_trusted_2021}&
  AIH 21 & \cite{houben_inspect_2022,winter_trusted_2021}  \\
 
 \hline
 AIH 4 & \cite{gebru_datasheets_2021,konigstorfer_ai_2022,raji_closing_2020} & 
 AIH 10 & \cite{mehrabi_survey_2021,steimers_sources_2022} &
 AIH 16&\cite{houben_inspect_2022,winter_trusted_2021}&
   AIH 22 & \cite{studer_towards_2021} \\
 
 \hline
 AIH 5 & \cite{bhatt_explainable_2020,steimers_sources_2022} & 
 AIH 11 & \cite{frank_effect_2017,willers_safety_2020} &
 AIH 17& \cite{dietterich_overfitting_1995,winter_trusted_2021} &
  AIH 23 & \cite{houben_inspect_2022,iso_5469_aisafety,steimers_sources_2022,willers_safety_2020} \\

 \hline
 AIH 6 & \cite{iso_5469_aisafety,steimers_sources_2022}& 
 AIH 12 & \cite{steimers_sources_2022,steinhardt_certified_nodate} &
 AIH 18 & \cite{bhatt_explainable_2020,houben_inspect_2022,iso_5469_aisafety,goos_establishing_2003} &
    AIH 24 & \cite{houben_inspect_2022,iso_5469_aisafety,willers_safety_2020}\\
 \hline

 \end{tabular}
 \end{table}
\section{AI Hazard Management}\label{sec: Met}
In this section, the AI Hazard Management framework is presented. It builds upon the preliminary AI hazard list shown in \Cref{sec: AI Hazard List} to fulfill its primary objectives, which are:

\textbf{1)} Systematically identify, assess, and treat all AI hazards.

\textbf{2)} Provide structured documentation to demonstrate that each identified AI hazard has either been mitigated to an acceptable level or the developer has been alerted to any shortcomings in the AI system caused by AI hazards.

\begin{figure}[h]
    \centering
    \includegraphics[scale = 0.42]{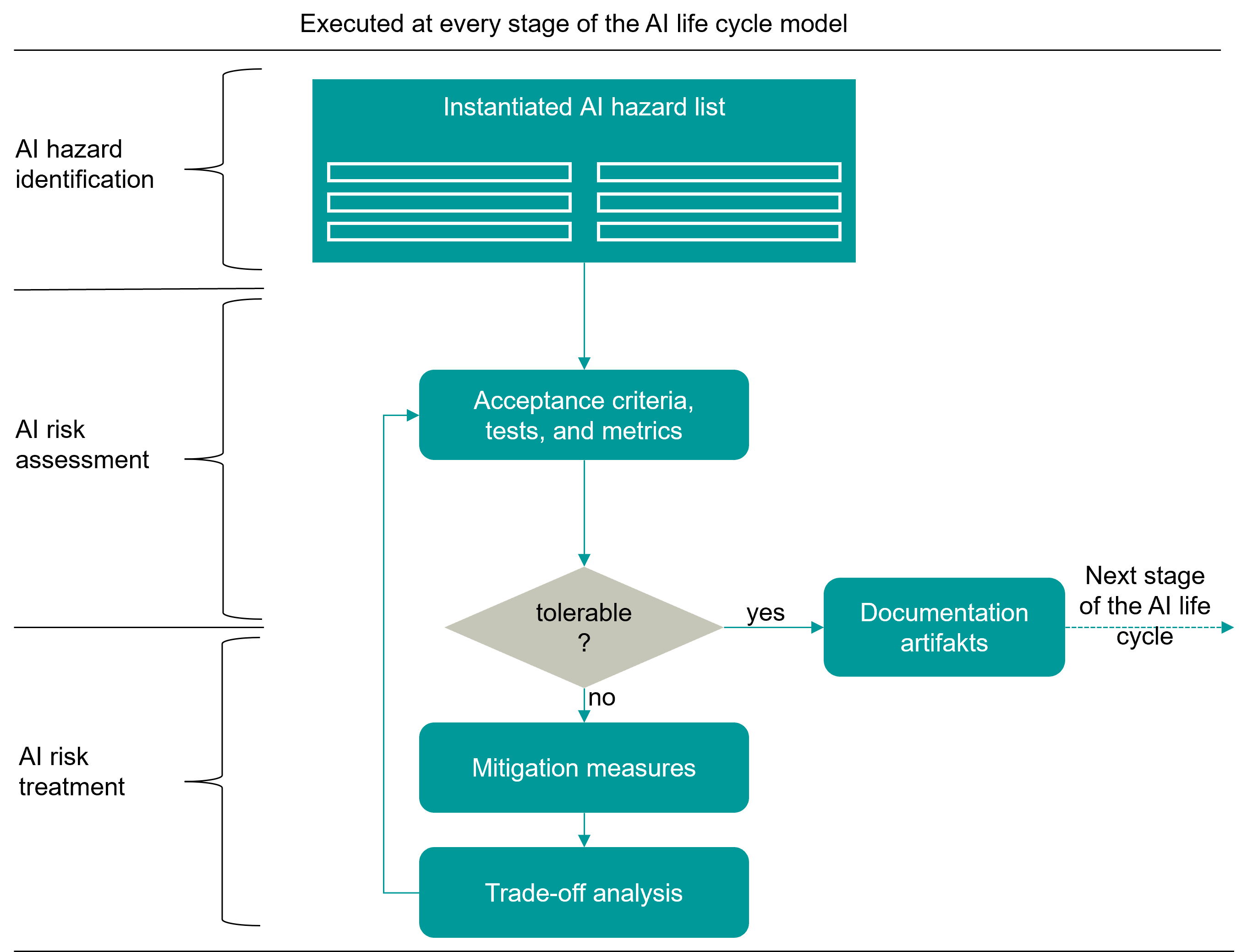}
    \caption{A schematic representation of the AIHM framework comprising the three main components: AI hazard identification, AI risk assessment, and AI risk treatment.}
    \label{fig: Process}
\end{figure}

With the AIHM framework, we aim to follow the general structure of a risk management process, such as given in the ISO 31000 \cite{iso_31000}, to ensure that it is compatible with existing risk management procedures in industry.
The four main components of the risk management process in ISO 31000 are: risk identification, risk estimation, risk evaluation, and risk treatment \cite{iso_31000}. 
We found that in AI development, the risk estimation and evaluation steps are highly overlapping, so we aggregated them into a single step called AI risk assessment. 
Therefore, the three steps of the AIHM process are:
1) AI hazard identification, 2) AI risk assessment, and 3) AI risk treatment.
The proposed process is depicted schematically in \Cref{fig: Process}, and details regarding the three steps are outlined in the remaining part of the section.
Additionally, as the transparency and auditability of the process are key features, continuous documentation of the process is required, which will be discussed at the end of the section. 
The AIHM process is executed at every stage of the AI life cycle. 

\subsection{AI hazard identification}\label{subsec: Identification}
The AI hazard identification is the initial step of the process and represents the risk identification as outlined in ISO 31000.
It aims to identify any possible AI hazard potentially causing AI risk.
In the AIHM, the outcome of the AI hazard identification step is an instantiated AI hazard list, which is the result of filtering the complete AI hazard list, as given in \Cref{sec: AI Hazard List}, in two steps:
Firstly, only AI hazards having the potential to occur at a particular stage of the AI life cycle are selected. This information can be obtained from the AI life cycle axis within the AI hazard list and taxonomy, see \Cref{sec: AI Hazard List}.
The second phase of the filtering process depends on the specific use case, as the impact of AI hazards may vary based on various factors, such as the AI technique (e.g., reinforcement learning vs. supervised learning) and the context of the AI application (e.g., human resources vs. industrial predictive maintenance). Therefore, this step is performed by a team consisting of domain experts and data scientists to ensure the required expertise. 
Reducing the instantiated AI hazard list to only the relevant AI hazards aims to increase the efficiency of the whole process.
Every decision regarding including or excluding AI hazards in the AIHC must be documented and justified with a clear argumentation.

\subsection{AI risk assessment}\label{subsec: Risk Assessment}
The AI risk assessment represents the risk estimation and evaluation steps outlined in ISO 31000. 
This step aims to assess the impact of AI hazards identified in the previous step and evaluate whether it is tolerable in a particular application.
To accomplish this, a test strategy has to be initialized, by determining appropriate metrics and corresponding acceptance criteria.
The exact execution of this step depends on the mode of the respective AI hazards. While no quantitative methods exist for procedural AI hazards, which rely on human errors, a possible form of assessment could be the provision of process documentation in which the involved party justifies all decisions and actions. A second person then evaluates this process documentation for correctness and meaningfulness. In the case of socio-technical or technical AI hazards, appropriate metrics, and acceptance criteria must be established, such as thresholds related to the chosen metrics.
The specification of the acceptance criteria also is highly dependent on the AI application's context and potential technical limitations. Therefore, having a team with a diverse background is beneficial for setting the acceptance criteria to cover all relevant aspects. This includes domain experts, data scientists, and business representatives.
For example, metrics such as false positive or negative rates can be utilized to evaluate the quality of a classification algorithm.
However, the consequences of a false negative classification in a medical decision-making system could be more severe than in a Human Resources application, classifying job applicants based on their suitability for an open position. Hence, the acceptance criteria for the metrics should also consider these different levels of consequences. 

\subsection{AI risk treatment}\label{subsec: Risk Mitgation}
The objective of the third step is to reduce the impact of AI hazards that have been evaluated as non-tolerable in the previous step. This corresponds with risk treatment in ISO 31000. 
Suitable AI risk mitigation measures shall be applied to reduce the impact of AI hazards. If several are available, a team of data scientists and domain experts should hold a council on the most suitable.
For example, a common problem in supervised learning is that a model might not be robust against small perturbations in the input space. Augmented training is a common way to mitigate this problem, which means re-training the supervised learning model with additional data gained by applying data augmentation techniques, e.g., \cite{wong_understanding_2016}.
 However, in some cases, the augmentation of training data is too complex, and the additional data sampling is too expensive. 
 In such a case, a different mitigation technique, for instance, using a different cost function (regularization), can be applied, see for instance \cite{moradi_survey_2020}.

Further problems in the AI risk treatment step are interactions and trade-offs between AI hazards. For example, although increasing the robustness of a model, regularization techniques might decrease the overall performance of a model, e.g., \cite{zhang_theoretically_2019}.
Consequently, acceptance criteria dependent on the overall performance can be violated if mitigation techniques improving the robustness of the model are applied. Therefore, such trade-offs should be considered in the decision process regarding suitable mitigation techniques within a trade-off analysis. Also, they need to be re-evaluated after applying a mitigation technique.
\subsection{Documentation}
The above-described process is iterated until all acceptance criteria are fulfilled or the completion of all acceptance criteria is evaluated as impossible.
In both cases, every step is to be documented in a structured form, including every action taken and argumentation for any decision made.
The developer must be notified if the risk cannot be reduced to an acceptable level. Consequently, additional safeguards shall be implemented. 
\section{Case Study}
\label{sec: Case Study}
In this section, we demonstrate how the AIHM framework provides guidance for risk management while developing an industrial AI application. As a case study, we examined a protective solution for power grids to address the challenge of detecting residual currents resulting from high-impedance ground faults (HIGF). These faults can pose significant dangers, such as wildfires, and are difficult to identify even with the expertise of domain specialists. The proposed solution utilizes a deep neural network (DNN) to classify the presence or absence of high-resistance faults in the distribution power grid by analyzing time series data of voltage and current. For simplicity, we assume the AI system to be a binary classifier, i.e. detecting the presence or non-presence of a HIGF in the grid. Note that the AI system is scoped as a decision support tool at the project's current state. This means that if it detects a HIGF, an alerting system informs a human user, who ultimately has the decision power over consequent actions.

The application of the AIHM to the power grid use case was simulated by conducting an interview with the lead data scientist and domain expert. This interview was performed in three sessions, taking a total of 4 hours. In this manner, decision-based steps in the AIHM could be simulated.
The execution of the AIHM framework is demonstrated by examining its implementation regarding four AI hazards: discriminative data bias, lack of robustness, inappropriate degree of transparency to end users, and data drift. These hazards are representative of various stages in the AI life cycle. With this, the different actions required to handle AI hazards are highlighted.
\\
\textbf{Discriminative data bias}\ 

\textit{AI risk identification}
Firstly, note that discriminative data bias relates to the data collection and preparation stage in the AI life cycle, see \Cref{tab: AI hazard overview}.
There are various types of how data bias can manifest itself in the data, including representation, sampling, aggregation, and omitted variable bias. For a comprehensive overview of these different types, we refer to \cite{mehrabi_survey_2021}.
The second filtering step described in \Cref{subsec: Identification} reviews the relevance of the AI hazard for the particular use case. The only data used in developing the AI system was time series data of voltage and current in the power grid. Therefore, arguably any impact of discriminative data bias could be ruled out. Hence, there is no need to consider this AI hazard in the subsequent assessment and mitigation steps.

\textit{AI risk assessment and treatment} no need for action

\textit{Documentation}
In summary, as the AI hazard was evaluated as irrelevant for this application, the impact of discriminative data bias is tolerable. Furthermore, this claim is supported by an argumentation regarding the data properties, ensuring required auditability. 
\\
\textbf{Lack of robustness}

\textit{AI risk identification}
The AI hazard lack of robustness is concerned in the modeling stage, see \Cref{sec: AI Hazard List}. The confrontation with adversarial data during operation is evaluated as extremely unlikely due to an attacker's extreme effort to create and infiltrate them into the AI system. Hence, robustness against adversarial attacks is not the developer's main concern. Since the application's operational data potentially suffers the influence of noisy measurements, the model's robustness against input perturbations is essential. 

\textit{AI risk assessment}
Consequently, suitable metrics and corresponding acceptance criteria need to be determined.
A possible way of assessing the robustness of a model is to evaluate its performance against meaningful perturbations in the input data. 
For example, there exist many techniques in the field of signal processing, see e.g. \cite{iwana_empirical_2021}
Concerning the data in this particular application, more sophisticated data augmentation is required. Data scientists and domain experts create an augmented data set, mainly using simulation-based augmentation techniques.
Regarding the acceptance criteria, a maximal tolerable performance decrease on this perturbed data set compared to the original one has to be determined. If this threshold is exceeded, the impact of lack of robustness is evaluated as non-tolerable.

\textit{AI risk treatment}
Assuming to observe a non-tolerable impact of this AI hazard, mitigation measures are required. For instance, the model could be re-trained using augmented data.
A possible negative side effect when increasing the robustness of a model is that this potentially causes a decrease in the model's overall performance. This is called the robustness-accuracy trade-off \cite{zhang_theoretically_2019}.
The model's overall performance is re-evaluated to ensure the mitigation technique has no adverse side effects.

\textit{AI risk assessment (revised)}
After re-training the model on the augmented training set, its estimation metric is also to be re-evaluated to see if the mitigation technique was sufficient. If evaluated as sufficient, the AI hazard is considered tolerable.

\textit{Documentation}
Every step and preliminary result, including the choice of estimation metrics, acceptance criteria, and evaluation results, shall be documented to ensure the auditability of the process.
\\
\textbf{Inappropriate degree of transparency to end users}

\textit{AI hazard identification}
The transparency of an AI system to the end user is concerned in the scoping stage as well as at deployment and operation. The AI system is based on a DNN and is considered a so-called black box model. This means that determining an output for a given input is not human-understandable, per se. 
However, since this decision-making has severe consequences (e.g., shutting down the power grid), more transparency is required. Hence, the AI hazard inappropriate degree of transparency to end users is identified as relevant and needs to be considered in the assessment and potentially in the mitigation step.

\textit{AI risk assessment}
Numerically quantifying transparency, in general, is not viable. Therefore, a qualitative estimation and evaluation might oppose a suitable solution. Determining what means are necessary to provide an appropriate degree of transparency depends on the application and user profile. In this application, the end-user are domain experts with a particular level of domain knowledge, which should be acknowledged in what kind of information the AI application needs to provide the user during operation. 

\textit{AI risk treatment}
To increase transparency during operation, a monitoring system is implemented. This includes a dashboard informing the user about activities made by the AI system. For instance, if the system predicts the presence of a HIGF in the grid, this is reported to the user, who can see the input data that led to this decision. By this means, a human is set in the loop.

\textit{Documentation}
Again every result is to be documented. 
\\
\textbf{Data drift}

\textit{AI risk identification}
Data drift describes the change of input distribution over time. If not managed properly, this issue can limit the performance of an AI system.
To understand whether this AI hazard is relevant for the power grid application, consider the following observations.
Electric grids differ in size and location. Especially with the latter, there come differences in climate-related factors such as temperature or humidity impacting the input data. Therefore, assuming to deploy the AI system in different locations, it is not guaranteed that it has equal performance. Also, factors such as climate change can influence the functionality quality of the AI system in the long run.
Therefore, data drift is relevant in this application.

\textit{AI risk assessment}
Data drift is an issue known to manifest over time after deployment. However, preventive activities can be executed during development. This includes considering foreseeable distribution shifts. For instance, in case of deploying the AI system in another power grid, the model can be adapted by re-training with data gained from the new location. 

\textit{AI risk reduction}
However, as not every shift in distribution can be certainly predicted, a periodic distribution shift detection mechanism can be implemented. This means the similarity between the training data and a sample collected in operation is evaluated in a pre-defined frequency. If a significant deviation in this distribution is detected, this indicates the presence of data drift. In such a case, countermeasures can be initialized, including re-training the model.

\textit{Documentation}
By laying out a concrete plan for handling data drift, requirements towards auditability can be fulfilled.

\section{Conclusion}\label{sec: Conclusion}
In this paper, we proposed the AI Hazard Management framework, which systematically manages AI risk's root causes. 
Further, it delivers evidence that their impact is reduced to a tolerable level.
The application to a power grid use case showed the framework's effectiveness in managing AI hazards.
The framework builds upon a preliminary list of AI hazards and a taxonomy supporting to manage the AI hazards.
An extension of the AI hazard list is subject to future work. Through the analysis of further AI use cases, we expect to identify further relevant pitfalls of AI. In addition, we plan to systematically investigate quantification and mitigation techniques for the list of identified AI hazards. 
\subsubsection{Acknowledgement}
This research has received funding from the Federal Ministry for Economic Affairs and Climate Action (BMWK) under grant agreements 19I21039A.

This preprint has not undergone peer review or any post-submission improvements or corrections. The Version of Record of this contribution is published in Frontiers of Artificial Intelligence, Ethics, and Multidisciplinary Applications: 1st International Conference on Frontiers of AI, Ethics, and Multidisciplinary Applications (FAIEMA), Greece, 2023, and is available online at https://doi.org/10.1007/978-981-99-9836-4\_27.

\bibliographystyle{splncs04}

\bibliography{AIHM.bib}

\end{document}